\documentclass{article}

\usepackage{microtype}
\usepackage{graphicx}
\usepackage{booktabs} 

\usepackage{amsmath,amsfonts,bm}

\newcommand{\RNum}[1]{\uppercase\expandafter{\romannumeral #1\relax}}
\newcommand{\Rnum}[1]{\lowercase\expandafter{\romannumeral #1\relax}}

\def\Secref#1{Section~\ref{#1}}
\def\Figref#1{Figure~\ref{#1}}
\def\Tabref#1{Table~\ref{#1}}

\def\Secref#1{Section~\ref{#1}}

\def\eqref#1{Eq. (\ref{#1})}

\def\Coref#1{Corollary~\ref{#1}}

\def\0{\bm{0}} 
\def\1{\bm{1}}

\def\vdelta{{\bm{\delta}}} 
\def\vtheta{{\bm{\theta}}}

\def\vg{{\bm{g}}}

\def\vv{{\bm{v}}}
\def\vw{{\bm{w}}}
\def\vx{{\bm{x}}}

\DeclareMathAlphabet{\mathsfit}{\encodingdefault}{\sfdefault}{m}{sl}
\SetMathAlphabet{\mathsfit}{bold}{\encodingdefault}{\sfdefault}{bx}{n}

\usepackage{subfig}
\usepackage{multirow}
\usepackage[table,xcdraw]{xcolor}
\usepackage{makecell}
\usepackage{hhline}

\usepackage{hyperref}

\usepackage[accepted]{arxiv_icml2023}

\usepackage{amsmath}
\usepackage{amssymb}
\usepackage{mathtools}
\usepackage{amsthm}

\usepackage[capitalize,noabbrev]{cleveref}

\theoremstyle{plain}
\newtheorem{theorem}{Theorem}[section]

\newtheorem{corollary}[theorem]{Corollary}
\theoremstyle{definition}

\theoremstyle{remark}

\usepackage[textsize=tiny]{todonotes}

\icmltitlerunning{Exploring the Effect of Multi-step Ascent in Sharpness-Aware Minimization}

\begin{document}

\twocolumn[
\icmltitle{Exploring the Effect of Multi-step Ascent in Sharpness-Aware Minimization}



\icmlsetsymbol{equal}{*}

\begin{icmlauthorlist}
\icmlauthor{Hoki Kim}{equal,snu}
\icmlauthor{Jinseong Park}{equal,snu}
\icmlauthor{Yujin Choi}{equal,snu}
\icmlauthor{Woojin Lee}{equal,dgu}
\icmlauthor{Jaewook Lee}{snu}
\end{icmlauthorlist}
\icmlaffiliation{snu}{Seoul National University, Seoul, Republic of Korea}
\icmlaffiliation{dgu}{Dongguk University-Seoul, Seoul, Republic of Korea}
\icmlcorrespondingauthor{Jaewook Lee}{jaewook@snu.ac.kr}

\icmlkeywords{Machine Learning, ICML}

\vskip 0.3in
]



\printAffiliationsAndNotice{\icmlEqualContribution} 

\begin{abstract}
Recently, Sharpness-Aware Minimization (SAM) has shown the state-of-the-art performance by seeking flat minima. 
To minimize the maximum loss within a neighborhood in the parameter space, SAM uses an ascent step, which perturbs the weights along the direction of gradient ascent with a given radius. 
While single-step or multi-step can be taken during ascent steps, previous studies have shown that multi-step ascent SAM rarely improves generalization performance. 
However, this phenomenon is particularly interesting because the multi-step ascent is expected to provide a better approximation of the maximum neighborhood loss. 
Therefore, in this paper, we analyze the effect of the number of ascent steps and investigate the difference between both single-step ascent SAM and multi-step ascent SAM. 
We identify the effect of the number of ascent on SAM optimization and reveal that single-step ascent SAM and multi-step ascent SAM exhibit distinct loss landscapes. 
Based on these observations, we finally suggest a simple modification that can mitigate the inefficiency of multi-step ascent SAM.
\end{abstract}

\section{Introduction}
\label{sec:Introduction}

\begin{figure}[t]
    \centering
    \includegraphics[width=0.85\linewidth]{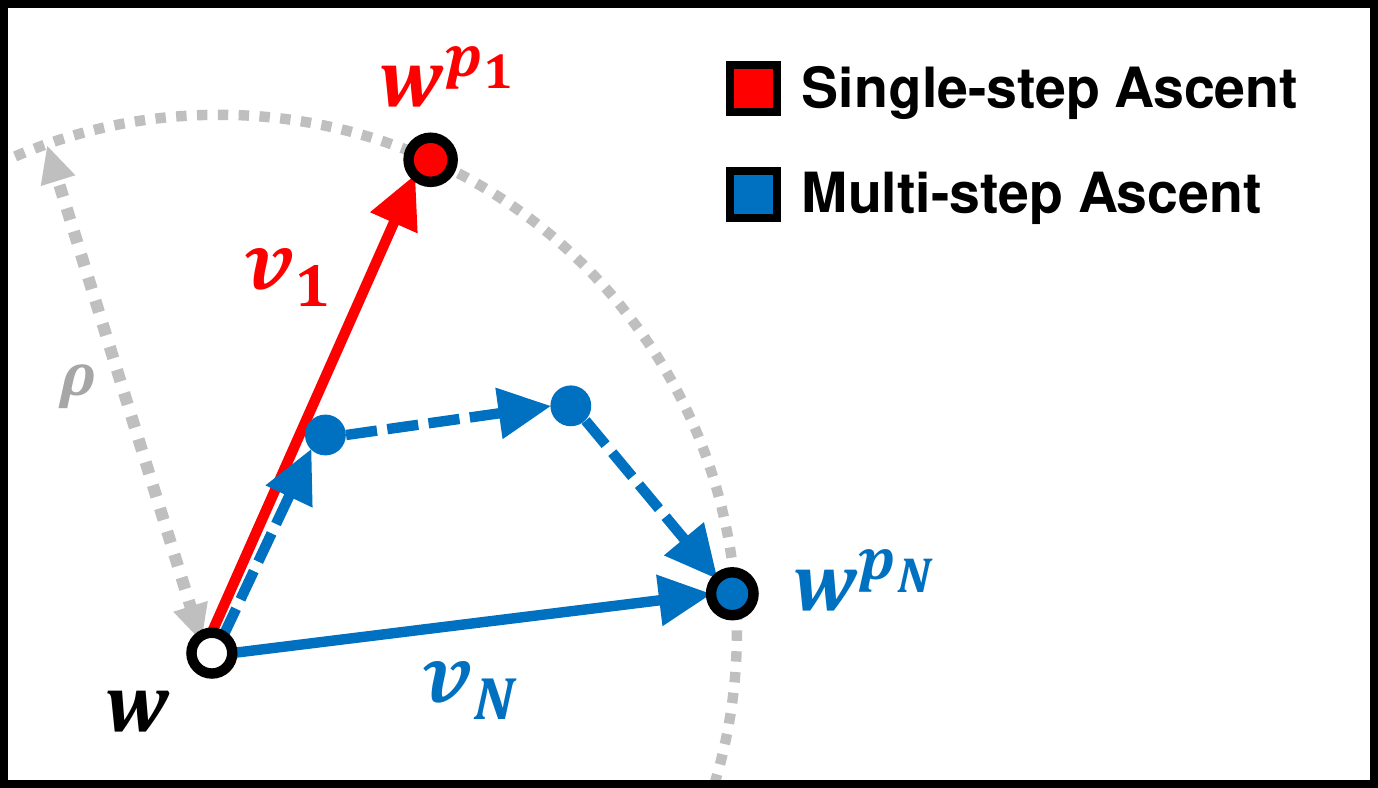}  
    \caption{Illustration of the single-step and multi-step ascent.}
    \label{fig:new_illustration}
\end{figure}

Deep neural networks are highly over-parameterized, allowing them to achieve near-zero training loss. However, this often results in poor generalization performance, as evidenced by the discrepancy between the performance of training data and test data \cite{neyshabur2017exploring, ishida2020we}. To understand and improve the generalization performance of neural networks, recent studies have explored the geometric properties of the loss landscape and have demonstrated that the sharpness of a minimum might be a core component of generalization \cite{mcallester1999pac, dinh2017sharp, keskar2017large, jiang2019fantastic}.

Among several sharpness-aware training methods \cite{izmailov2018averaging, he2019asymmetric}, Sharpness-Aware Minimization (SAM) \cite{foret2020sharpness} has shown state-of-the-art generalization performance. SAM aims to reach flat minima by minimizing the maximum loss within a neighborhood in the parameter space. 
To solve this min-max problem, SAM uses an \textit{ascent step} that perturbs the weight along the loss increasing direction. 

\begin{figure*}[t!]
    \centering
    \subfloat[CIFAR-10 \label{fig:Barchart_CIFAR10}]{%
       \includegraphics[width=0.47\linewidth]{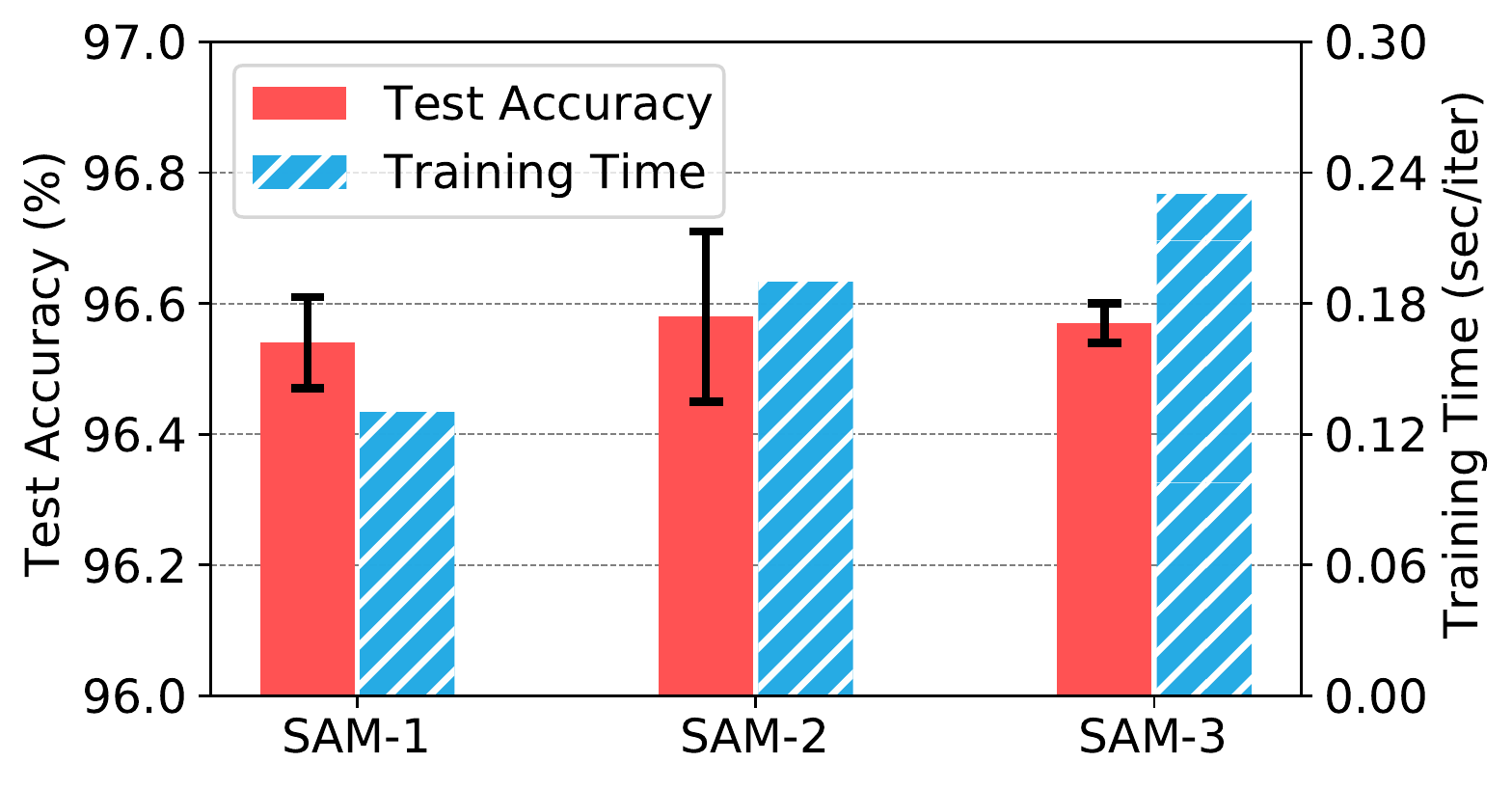}}
    \subfloat[CIFAR-100 \label{fig:Barchart_CIFAR100}]{%
       \includegraphics[width=0.47\linewidth]{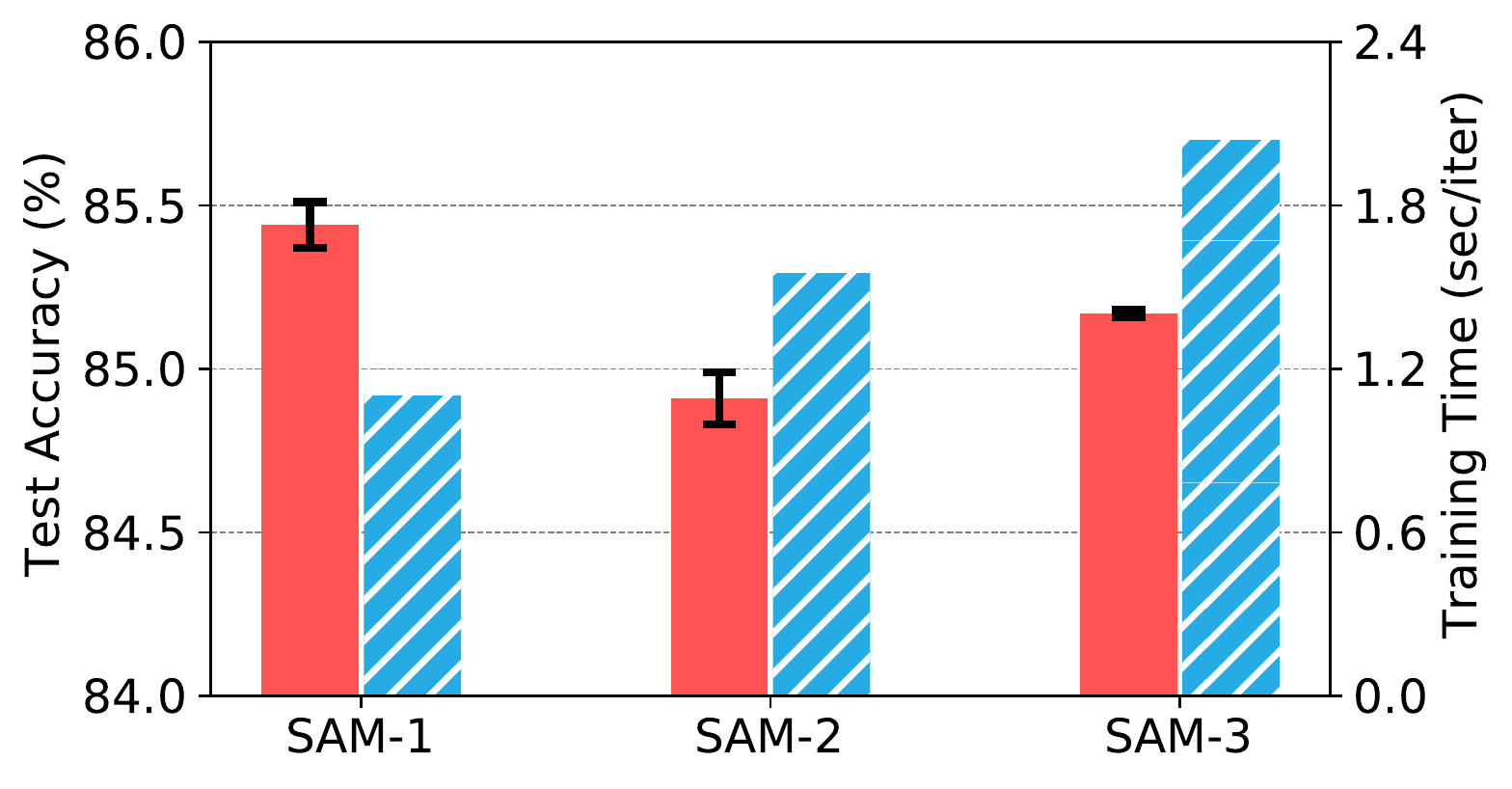}}
    \caption{Test accuracy and training time for SAM-$1$, SAM-$2$, and SAM-$3$, which use $N=1$, $2$, and $3$ ascent steps during training, respectively. For both datasets, SAM-$2$ and SAM-$3$ require more time but do NOT guarantee better performance than SAM-$1$.}
    \label{fig:Barchart}
\end{figure*}

As the ascent step, two different methods can be taken: \textit{single-step ascent} and \textit{multi-step ascent}. We provide a simple illustration of single-step ascent and multi-step ascent in \Figref{fig:new_illustration}. The single-step ascent perturbs the weight along the direction of gradient ascent only once, whereas the multi-step ascent uses several iterations to perturb the weight \cite{foret2020sharpness, wu2020adversarial}. Considering that the loss landscape of deep learning models is often highly non-linear \cite{du2017gradient, pmlr-v80-kleinberg18a}, the multi-step ascent is expected to yield a more precise solution as has been observed in similar min-max problems \cite{madry2018towards, wong2020fast}. 

However, contrary to this expectation, prior studies have argued that using the multi-step ascent during SAM training does not guarantee better performance than the single-step ascent SAM \cite{foret2020sharpness, wu2020adversarial}. 
Let us denote SAM using $N$ ascent steps as SAM-$N$. As shown in \Figref{fig:Barchart}, SAM-$2$ and SAM-$3$ show similar or even worse generalization performance compared to SAM-$1$, despite the increased computational cost. Although this inefficiency of multi-step ascent SAM is contrary to the common belief, however, it still remains largely unexplored.

Thus, in this paper, we aim to reveal the reasons for the inefficiency of multi-step ascent SAM. In \Secref{sec:revisit}, we first analyze how the number of ascent steps affects the perturbed weights and their gradients during SAM optimization. In \Secref{sec:sm}, we identify the distinct loss landscapes of SAM-$1$ and SAM-$N$ and suggest a simple modification that mitigates the inefficiency of multi-step ascent SAM. In \Secref{sec:experiments}, we demonstrate the effectiveness of the proposed modification through experiments on various models and datasets, including CIFAR-10, CIFAR-100, and ImageNet.

Our main contributions can be summarized as follows:
\begin{itemize}
    \item We conduct a comprehensive analysis of the number of ascent steps and identify their effect on perturbed weights and gradients.
    \item  We reveal that single-step and multi-step ascent SAM have distinct loss landscapes with respect to the number of ascent steps.
    \item We suggest a simple modification to mitigate the inefficiency of multi-step ascent SAM and demonstrate that it can provide better generalization performance.
\end{itemize}

\section{Background and Related Work}
\subsection{Sharpness-Aware Minimization}
The key idea of SAM is to minimize the maximum loss in the vicinity of a weight $\vw$ within a radius $\rho$:
\begin{equation}
    \min_\vw \max_{\|\vv\|=1} \ell(\vw+\rho\vv).
    \label{eq:sam}
\end{equation}
To solve this min-max optimization problem, SAM utilizes two sequential steps: an \textit{ascent step} and a \textit{descent step}.

In the ascent step, SAM first calculates the perturbed weight, which we denote $\vw^{p}$, as follows:
\begin{align}
    \text{\underline{Ascent step}:}& \quad \vw^{p}_t = \vw_t + \rho \vv,
    \label{eq:ascent}
\end{align}
where $\vv$ is an ascent direction.

\textbf{Single-step ascent.}
To calculate the ascent direction $\vv$, \citet{foret2020sharpness} typically used the first-order Taylor approximation as follows:
\begin{align}
    \vw^{p_1} = \vw + \rho \cdot \vv_1 \textit{\quad where\quad} \vv_1 = \frac{\nabla \ell(\vw)}{\|\nabla \ell(\vw)\|},
    \label{eq:single-step_ascent}
\end{align}
where $\nabla \ell(\vtheta) := \nabla_\vtheta \ell(\vtheta)$ for any $\vtheta$ unless otherwise specified.

Under the assumption of linearity, this \textit{single-step ascent} yields the maximum loss in its $\rho$-neighborhood in the parameter space. However, since the loss function of deep learning models is generally non-linear, there is no guarantee that single-step ascent will always result in the maximum loss.

\textbf{Multi-step ascent.}
To alleviate the limitations of linear approximation, \citet{foret2020sharpness} also explored \textit{multi-step ascent} with the number of steps $N$. For each iteration for $n \in \{1, ..., N\}$, the following is performed:
\begin{align}
    \vw^{p^{n}_N} &= \vw^{p^{n-1}_N} + \rho_n \cdot \frac{\nabla \ell(\vw^{p^{n-1}_N})}{\|\nabla \ell(\vw^{p^{n-1}_N})\|}, \label{eq:multi_ascent}
\end{align}
where $\vw^{p^0_N}=\vw$ and $\rho_n$ is the step-size at the $n$-th iteration.
Unless specified otherwise, $\rho_n=\rho/N$. Given the above equation, the ascent direction can be formalized as:
\begin{align}
    \vv_N &= \frac{\vw^{p^N_N} - \vw}{\| \vw^{p^N_N} - \vw \|},
    \label{eq:multi_ascent_direction}
\end{align}
and thus the final perturbed weight becomes 
\begin{align}
    \vw^{p_N}=\vw+\rho\vv_N.
    \label{eq:perturbed_weight}
\end{align}

After calculating the perturbed weight $\vw^p$, SAM calculates a \textit{perturbed loss} $\ell(\vw^{p})$ and uses its gradient $\nabla \ell(\vw^{p})$ to update the weight $\vw$ as follows:
\begin{align}
    \text{\underline{Descent step}:}& \quad \vw_{t+1} = \vw_t - \eta \nabla \ell(\vw^{p}_t).
    \label{eq:descent}
\end{align}
where $\eta$ is the learning rate.

\subsection{Understanding Mechanisms of SAM}
In light of the success of SAM, several studies have been conducted to further comprehend and enhance its performance: suggesting a transformation technique of parameters to achieve scale-invariant sharpness \cite{kwon2021asam}, alleviating certain limitations in optimization by subtracting $\nabla \ell (\vw)$ from $\nabla \ell (\vw^p)$ \cite{zhuang2021surrogate}, exploring the effectiveness of a random perturbation during weight perturbation \cite{liu2022random}, and reducing the computational cost during SAM training \cite{du2021efficient}.

However, the majority of these works have primarily focused on single-step ascent SAM. Only a few studies, such as \cite{andriushchenko2022towards}, have explored the multi-step ascent, but have confirmed that utilizing multiple steps rarely improves the performance of SAM. Despite this, there is a lack of studies on the effect of the number of ascent steps. Thus, we aim to investigate the effect of the number of ascent steps to gain a better understanding of this phenomenon and the mechanism of SAM.

\subsection{Analyzing Loss Landscape} \label{subsec:analyzing_loss}

The analysis of loss landscapes has been a critical component in comprehending deep learning models, as it offers insight into the fundamental geometry of the parameter space \cite{du2017gradient, li2018visualizing}.
Indeed, analyzing the loss landscape has provided a deeper understanding of the properties of deep learning models in various domains \cite{kim2021understanding, lewkowycz2020large}.

As the ascent step aims to approximate the maximum loss in the loss landscape, we believe that investigating loss landscapes might be a key component for identifying the difference between SAM-$1$ and SAM-$N$ and understanding the mechanism of SAM. Indeed, it has been observed that the shape of the loss landscape significantly influences the convergence of SAM \cite{zhuang2021surrogate, kim2023stability}. observed that depends on the shape of the loss landscape. Thus, in this work, we measure the effect of the number of ascent steps by analyzing the loss landscape during and compare the loss landscapes of SAM-$1$ and SAM-$N$.

\section{Revisiting the Number of Ascent Steps}\label{sec:revisit}
In this section, we provide a step-by-step analysis on the effect of the number of ascent steps $N$ on each iteration of SAM optimization as follows:
\begin{equation*}
    {\underbrace{\vw \leftarrow \vw - \eta \overbrace{\nabla \underbrace{\ell(\vw^{p_N})}_{\text{Sec.3.1}}}^{\text{Sec.3.2}}}_{\text{Sec.3.3}}}
\end{equation*}
In \Secref{sec:inner}, we investigate the loss of the perturbed weights with respect to the number of ascent steps, $\ell(\vw^{p_N})$. In \Secref{sec:middle}, we analyze the gradients of the perturbed weights, $\nabla \ell(\vw^{p_N})$. In \Secref{sec:outer}, we finally investigate the effect of their gradients on the loss landscape by the update process, $\vw - \eta \nabla \ell(\vw^{p_N})$. To provide an easier understanding, we fix the model to be analyzed as ResNet-18 trained on CIFAR-10 with $\rho=0.1$. Nevertheless, similar results were observed for other settings regardless of the number of training epochs, training methods, and datasets. Detailed settings are presented in Appendix.

\subsection{Higher Number of Ascent Steps Yields Perturbed Weight with Higher Loss}\label{sec:inner}

As the initial step of SAM optimization, the perturbed weight $\vw^p$ is calculated to approximate the maximum loss in its vicinity. Thus, we first measure the effect of the number of ascent steps on the perturbed loss, $\ell(\vw^{p_N})$. Given the model, we calculate the ascent direction $\vv_N$ according to \eqref{eq:multi_ascent_direction} on the first mini-batch of the training data with varying $N$. Then, we plot the loss increase with respect to each $\vv_N$ for the same mini-batch.

\begin{figure}[t]
    \centering
    \includegraphics[width=0.9\linewidth]{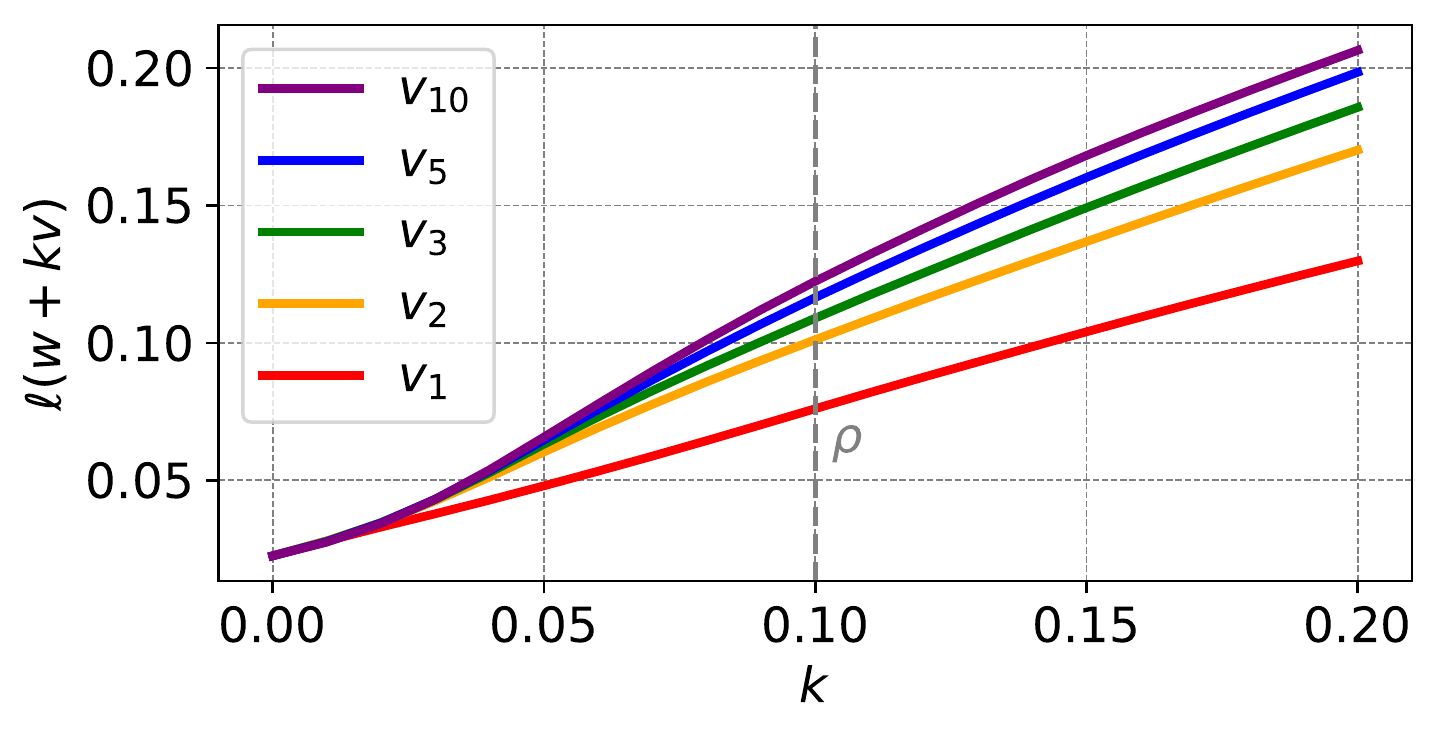}  
    \caption{Loss increase along the direction of $\vv_N$ with different number of ascent steps $N$. Perturbed loss $\ell(\vw+k\vv_N)$ increases as $N$ increases.}
    \label{fig:rho_sam}
\end{figure}

\Figref{fig:rho_sam} shows the loss increase along the ascent directions $\vv_N$ with varying $N$.
When we only consider $\vv_1$, which is a commonly used direction for measuring perturbed loss and sharpness \cite{foret2020sharpness, zhuang2021surrogate}, it may seem that the loss landscape is locally linear.
However, the loss value increases as $N$ increases for all $k$. Specifically, the perturbed loss with $\vv_{10}$ is significantly higher than that of $\vv_1$. Thus, we confirm that the linearity assumption on the loss landscape cannot be satisfied.

\begin{figure}[t]
    \centering
    \vspace{-10pt}
    \includegraphics[width=0.75\linewidth]{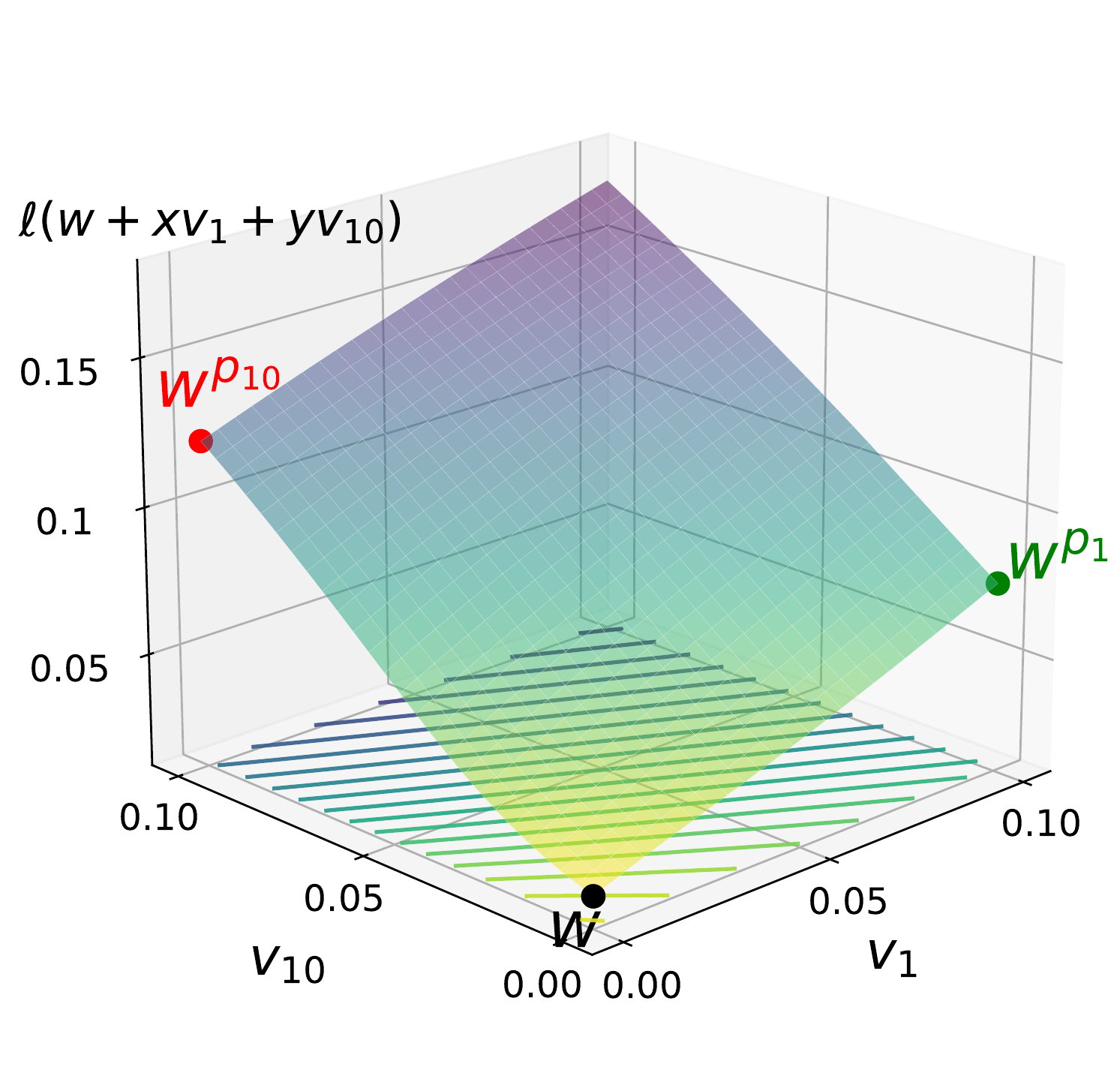}  
    \caption{Loss landscape interpolated by $\vv_1$ with $\vv_{10}$. A higher number of ascent steps yields a higher perturbed loss than any other combination of ascent directions.}
    \label{fig:3d_plot_final}
\end{figure}

In \Figref{fig:3d_plot_final}, we also plot the loss landscape with respect to the linear combination of two different ascent directions with varying $N$. First, we can observe that the loss landscape in the parameter space with respect to ascent directions is not highly distorted. Most importantly, The direction that generates the highest loss increase is $\vv_{10}$, rather than $\vv_1$ or any linear combination of $\vv_1$ and $\vv_{10}$. We observe the same results for all $\vv_N$ with $N\geq2$. 

Thus, from the above results, we can conclude that the loss landscape in the parameter space is non-linear, and thus the multi-step ascent generally yields a higher perturbed loss $\ell(\vw^p)$  than the single-step ascent.

\subsection{Gradient Difference between Perturbed Weights}\label{sec:middle}

Given the fact that different numbers of ascent steps yield different perturbed losses, we now focus on the gradients of the perturbed weight $\nabla\ell(\vw^{p_N})$. Since the descent step in \eqref{eq:descent} is conducted by the gradient of the perturbed weight, $\nabla\ell(\vw^{p})$ is a crucial component of the optimization process.
In order to measure the effect of the number of ascent steps on the gradients, we analyze the cosine similarity between the gradients obtained by varying the number of ascent steps. Given the model and the perturbed weights $\vw^{p_N}$, we gather the gradients $\nabla \ell(\vw^{p_N})$ and measure their cosine similarities.

\begin{figure}[t]
    \centering
    \includegraphics[width=0.65\linewidth]{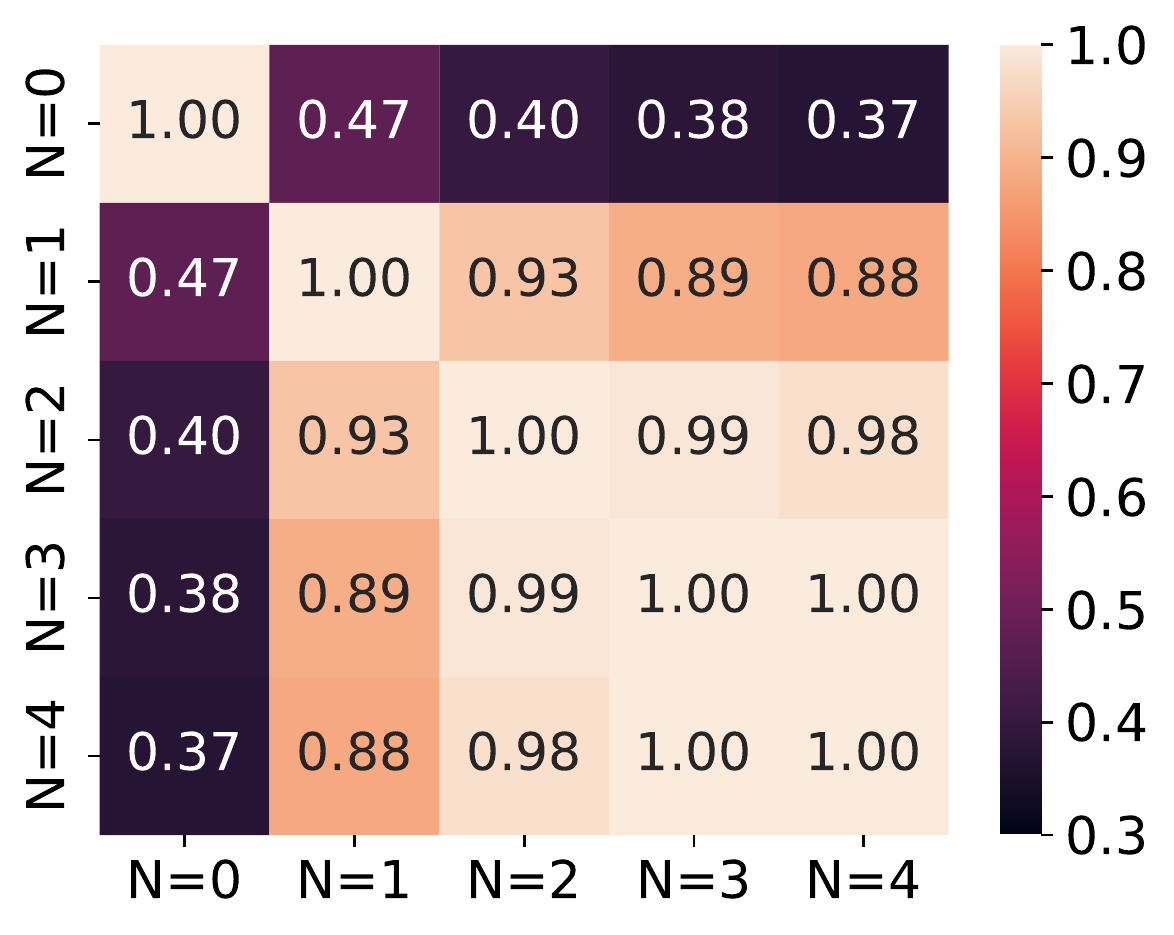}
    \caption{Cosine similarities of the gradients of the perturbed loss for various ascent steps $N$, i.e., $\cos(\nabla\ell(\vw^{p_N}), \nabla\ell(\vw^{p_N}))$.}
    \label{fig:cosine}
\end{figure}

\Figref{fig:cosine} shows the cosine similarities of the gradients for various ascent steps $N$. Here, $\nabla\ell(\vw^{p_0})=\nabla\ell(\vw)$. It is evident that the gradients of the vanilla gradient descent and SAM are easily distinguishable. The large difference between $\nabla\ell(\vw)$ and $\nabla\ell(\vw^{p_1})$ can be related to the performance gap between vanilla training and SAM, which highlights the importance of the direction $\nabla\ell(\vw^{p_1})$ as described in prior studies \cite{foret2020sharpness, zhuang2021surrogate}.

\begin{figure}[t]
    \centering
    \includegraphics[width=0.6\linewidth]{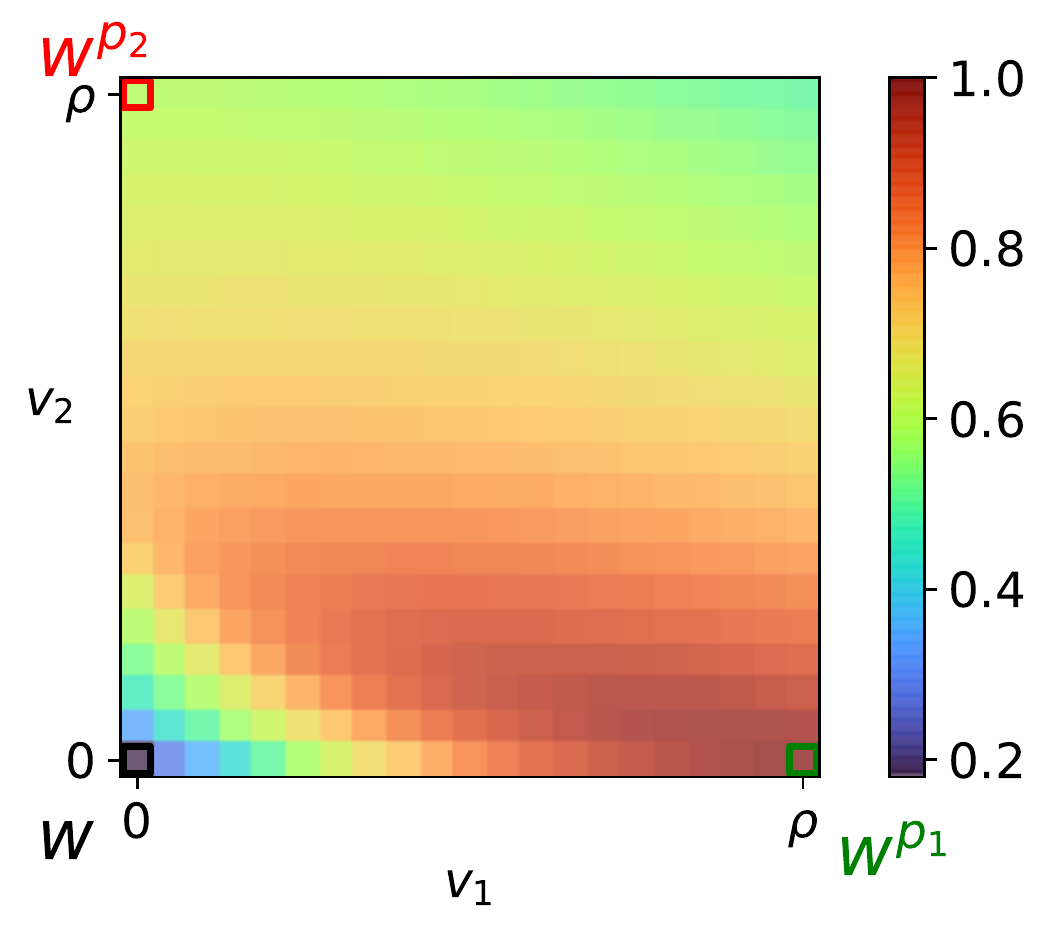}  
    \caption{Cosine similarity between $\nabla\ell(\vw^{p_1})$ and gradients of other perturbed weights in the parameter space spanned by $\vv_1$ and $\vv_2$, i.e., $\nabla\ell(\vw+x\vv_1+y\vv_2)$ with a larger $\rho=0.5$ during ascent steps. The difference in gradients is more pronounced when wider parameter space is considered.}
    \label{fig:weight_space_grad5}
\end{figure}

Most importantly, there is a difference in the cosine similarities between the gradient with $N=1, 2, 3$, and $4$. As the number of ascent steps increases, the cosine similarity between $\nabla\ell(\vw)$ and $\nabla \ell(\vw^{p_{N}})$ decreases. For example, when $N=1$, the cosine similarity is 0.47, but it decreases to 0.37 when $N=4$.
Additionally, as $N$ increases, the cosine similarity between $\nabla\ell(\vw^{p_1})$ and $\nabla \ell(\vw^{p_{N}})$ also decreases. For instance, the cosine similarity between $\nabla\ell(\vw^{p_1})$ and $\nabla \ell(\vw^{p_{2}})$ is 0.93, while the cosine similarity between $\nabla\ell(\vw^{p_1})$ and $\nabla \ell(\vw^{p_{4}})$ is 0.88. 

This difference between gradients can be even more pronounced when wider parameter space is considered. In \Figref{fig:weight_space_grad5}, we measure the cosine similarity with a radius of $\rho=0.5$, which is five times larger than the value used in \Figref{fig:cosine}. The cosine similarity between $\nabla\ell(\vw^{p_1})$ and $\nabla \ell(\vw^{p_{2}})$ is only approximately 0.6 (upper left cell), which is much lower than 0.88 in \Figref{fig:cosine}. Thus, the above results indicate that varying ascent steps not only results in a different perturbed weight, but also a different gradient.

\subsection{Distinct Effects of Gradients on Loss Decrease}\label{sec:outer}

Until this point, we have observed that the perturbed weights and their gradients vary depending on the number of ascent steps. Now, we finally investigate their effect on minimizing the perturbed loss. To evaluate this effect, we measure the actual decrease in loss, which can be estimated as follows:
\begin{equation}
    \Delta\ell(\vw) := \ell(\vw) - \ell(\vw - \eta \vg),
\end{equation}
where $\eta$ is a learning rate and $\vg$ is an update gradient. We measure the loss decreases for each gradient $\nabla\ell(\vw^{p_N})$ for different $N$ with the learning rate $0.1$ on the first training batch.

\begin{figure}[t]
    \centering
    \includegraphics[width=1\linewidth]{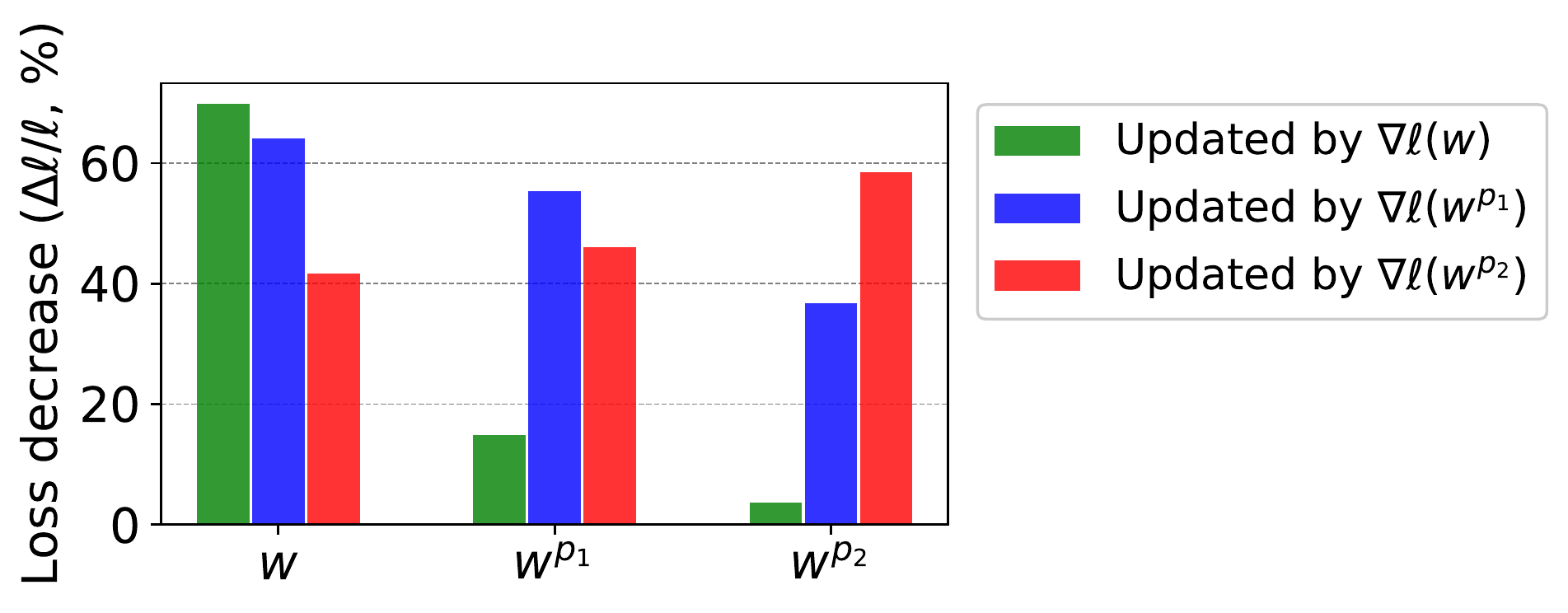}
    \caption{Loss decrease of each weight for each gradient $\nabla\ell(\vw^{p_N})$ with varying $N$.}
    \label{fig:loss_decrease_all}
\end{figure}

As shown in \Figref{fig:loss_decrease_all}, each gradient exhibits a distinct effect on each weight. $\nabla\ell(\vw)$ shows the most significant loss decrease of $\vw$, yet it does not effectively minimize the perturbed losses of $\vw^{p_1}$ and $\vw^{p_2}$ (green). This is consistent with the findings of \cite{foret2020sharpness, zhuang2021surrogate} that the gradient of SGD has difficulty in minimizing the perturbed loss. 
In contrast, the gradients of perturbed weights lead to a significant decrease in perturbed losses. However, as the number of ascent steps increases, the loss decrease of $\vw$ diminishes. This can be related to \Figref{fig:cosine}, which shows a low cosine similarity between $\nabla\ell(\vw)$ and $\nabla\ell(\vw^{p_N})$ as $N$ increases. In other words, as $N$ increases, it is able to effectively minimize its perturbed loss $\ell(\vw^{p_N})$, but has difficulty in minimizing the loss of the current weight $\ell(\vw)$.

\begin{figure}[t]
    \centering
    \includegraphics[width=0.85\linewidth]{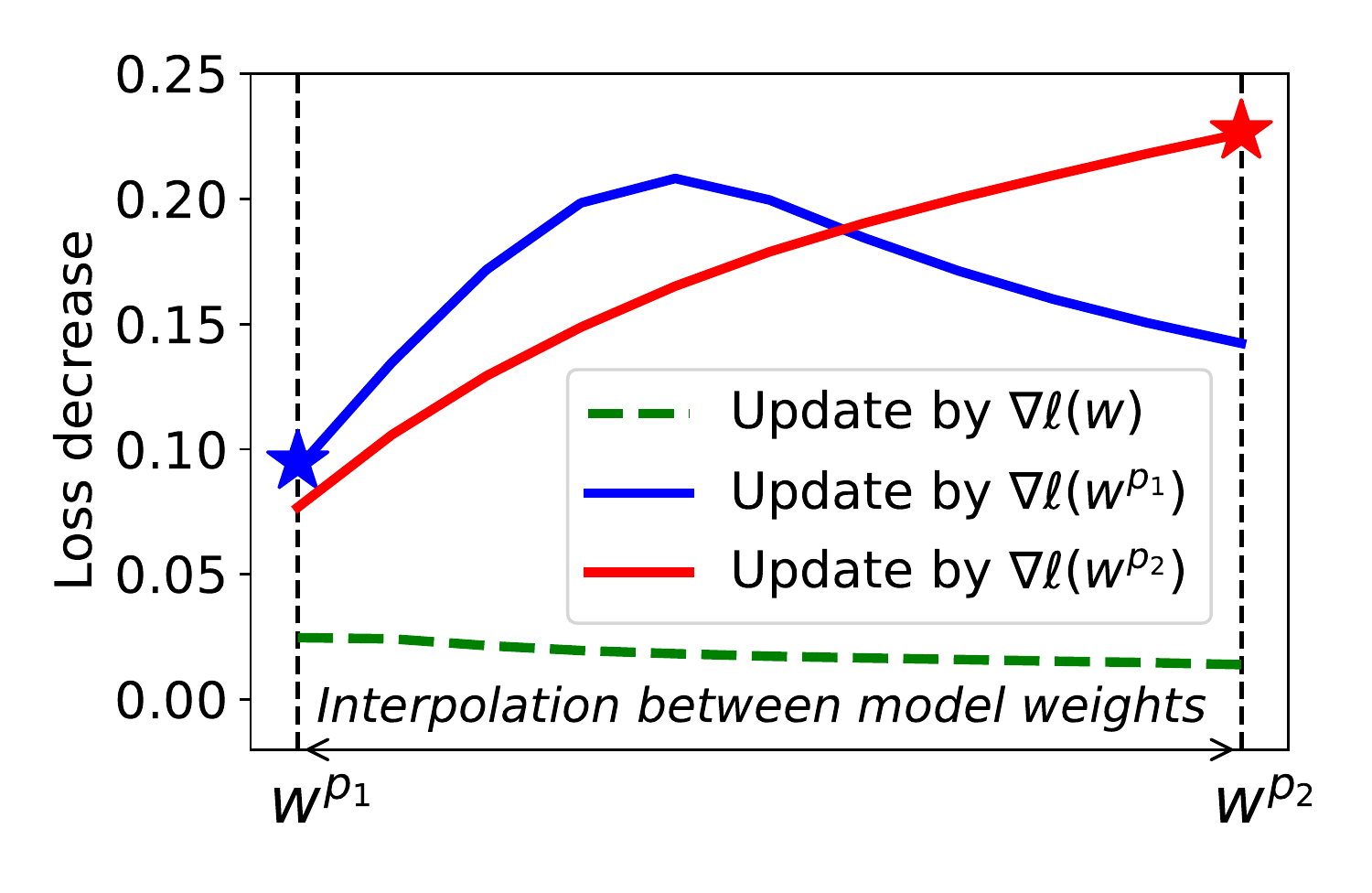}
    \caption{Loss decrease of interpolated perturbed weights for each gradient $\nabla\ell(\vw^{p_N})$ with varying $N$.}
    \label{fig:interpolated}
\end{figure}

In \Figref{fig:interpolated}, we further measure the loss decreases of the linearly interpolated perturbed weights between $\vw^{p_1}$ and $\vw^{p_2}$. While the gradient of the current weight show less effectiveness in minimizing the perturbed losses, the gradients of each perturbed weight show high loss decreases in the vicinity of their own weights. Specifically, as the blue and red star show, the perturbed losses are efficiently minimized by their own gradients.

In conclusion, varying the number of ascent steps yields distinct gradient effects on the loss landscape for each iteration, which could lead to different convergence of SAM optimization.

\section{Single-step versus Multi-step Ascent SAM}\label{sec:sm}
\begin{table*}[ht!] %
\centering
\caption{Analyzing on the loss landscape of SAM-$1$, SAM-$2$, and SAM-$3$. The number with the bold and green background denotes the minimum loss for each column. Different loss landscapes are observed between SAM-$N$.}
\label{tab:loss_per_steps}
\resizebox{.9\textwidth}{!}{%
\setlength{\tabcolsep}{1.5em}
\begin{tabular}{c|c|r|rrr}
\Xhline{3\arrayrulewidth}
\multicolumn{1}{c|}{\multirow{2}{*}{\textbf{Datasets}}}                                & \multirow{2}{*}{\begin{tabular}[c]{@{}c@{}} \textbf{Training Method} \end{tabular}} & 
 \multicolumn{1}{c|}{\multirow{2}{*}{\textbf{Loss $\ell(\vw)$}}}   
 & \multicolumn{3}{c}{\textbf{Perturbed loss } $\ell(\vw^{p_N}):=\ell(\vw+\rho\vv_N)$}                                    \\
\multicolumn{1}{l|}{}                                                 &                                                                                  &       & \multicolumn{1}{c}{$N=1$}       & \multicolumn{1}{c}{$N=3$}       & \multicolumn{1}{c}{$N=5$}       \\ \hline
\multirow{3}{*}{\begin{tabular}[c]{@{}c@{}}CIFAR-10\end{tabular}}  & SAM-$1$              & \cellcolor[HTML]{E5FFE5}\textbf{0.018\tiny{±0.002}} & 0.097\tiny{±0.004}          & 0.125\tiny{±0.005}          & 0.130\tiny{±0.005}          \\
                                                                      & SAM-$2$                                                                               & 0.019\tiny{±0.000}          & \cellcolor[HTML]{E5FFE5}\textbf{0.095\tiny{±0.001}} & \cellcolor[HTML]{E5FFE5}\textbf{0.119\tiny{±0.001}} & 0.124\tiny{±0.001}          \\
                                                                      & SAM-$3$                                                      & 0.021\tiny{±0.001}          & \cellcolor[HTML]{E5FFE5}\textbf{0.095\tiny{±0.004}} & \cellcolor[HTML]{E5FFE5}\textbf{0.119\tiny{±0.005}} & \cellcolor[HTML]{E5FFE5}\textbf{0.123\tiny{±0.005}} \\ \hline
\multirow{3}{*}{\begin{tabular}[c]{@{}c@{}}CIFAR-100\end{tabular}} & SAM-$1$                                                                               & \cellcolor[HTML]{E5FFE5}\textbf{0.166\tiny{±0.008}} & \cellcolor[HTML]{E5FFE5}\textbf{0.744\tiny{±0.018}}          & 1.285\tiny{±0.022}          & 1.416\tiny{±0.021}          \\
                                                                      & SAM-$2$                                                                             & 0.226\tiny{±0.008}          & 0.810\tiny{±0.007} & 1.198\tiny{±0.026}          & 1.280\tiny{±0.030}          \\
                                                                      & SAM-$3$                                                                               & 0.251\tiny{±0.004}          & 0.816\tiny{±0.019}          & \cellcolor[HTML]{E5FFE5}\textbf{1.180\tiny{±0.018}} & \cellcolor[HTML]{E5FFE5}\textbf{1.262\tiny{±0.027}}

\\\Xhline{3\arrayrulewidth}
\end{tabular}%
}
\end{table*}

In the previous section, we analyzed the effect of the number of ascent steps on each optimization step and demonstrated its substantial effect on the perturbed weights, gradients, and loss landscapes. Inspired by the observations, in this section, we investigate the difference between the final models at the end of the epoch trained with different numbers of ascent steps, i.e., the single-step ascent SAM (SAM-$1$) and multi-step ascent SAM (SAM-$N$), beyond their performance. Then, based on the observations, we suggest a simple modification that mitigates the inefficiency of multi-step ascent SAM.

\subsection{Distinct Loss Landscape of SAM-$1$ and SAM-$N$}\label{sec:distinct}

First, we would like to remark that SAM-$1$ and SAM-$N$ appear to be similar in terms of performance in \Figref{fig:Barchart}.
However, given that the number of ascent steps has a significant impact on the optimization of SAM, we hypothesize that SAM-$1$ and SAM-$N$ have distinct characteristics beyond their performance. Therefore, we compare the loss landscape of three different models, namely SAM-$1$, SAM-$2$, and SAM-$3$ in terms of their loss landscapes. Note that these models are the same as those presented in \Figref{fig:Barchart}, which summarizes their performance and computational costs.

In \Tabref{tab:loss_per_steps}, we measure the average and standard deviation of three different perturbed losses for each model for different numbers of ascent steps $N=1,3$, and $5$, over three different random seeds. More than five steps in ascent steps do not significantly increase the perturbed loss. The evaluation is conducted on the entire training set, using the same batch-size. Interestingly, models trained with different numbers of ascent steps exhibit different loss landscapes with respect to ascent directions. Specifically, SAM-$1$ has the lowest loss $\ell(\vw)$, but shows the highest perturbed loss $\ell(\vw^{p_5})$. In contrast, SAM-$2$ and SAM-$3$ have a lower perturbed loss $\ell(\vw^{p_5})$ than SAM-$1$, but exhibit a higher loss $\ell(\vw)$ than SAM-$1$ for all cases.

\begin{figure}[t]
    \centering
    \includegraphics[width=0.95\linewidth]{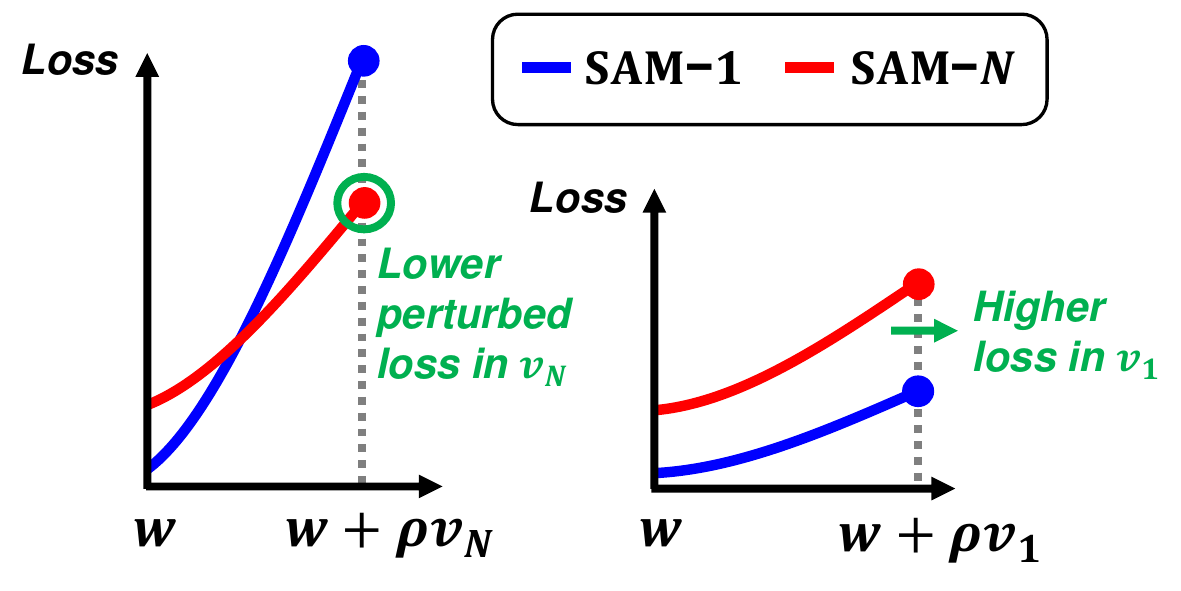}
    \caption{Illustration for distinct loss landscapes of SAM-$1$ and SAM-$N$. (Left) SAM-$N$ achieves a lower perturbed loss than SAM-$1$ in the direction of $\vv_N$; (Right) however, when considering a single-step ascent $\vv_1$, SAM-$N$ has a higher loss than SAM-$1$.}
    \label{fig:loss_diff}
\end{figure}

This result can be illustrated in \Figref{fig:loss_diff}. SAM-$N$ successfully minimizes a higher perturbed loss as it achieves a lower perturbed loss in the direction of $\vv_N$ compared to SAM-$1$ (e.g., left plot of \Figref{fig:loss_diff}). However, it fails to minimize losses in the direction of $\vv_1$ compared to SAM-$1$ (e.g., right plot of \Figref{fig:loss_diff}).
In other words, SAM-$N$ is more powerful in minimizing a higher perturbed loss, but it has difficulty in minimizing the loss of current weight and perturbed weights in ascent directions with a lower number of steps. 
Notably, this phenomenon is consistent with the observations in \Figref{fig:cosine} and \Figref{fig:loss_decrease_all}, which demonstrate $\nabla\ell(\vw^{p_N})$ can have difficulty in minimizing $\ell(\vw)$ and $\ell(\vw^{p_1})$ for $N\geq 2$. In terms of achieving the ideal flatness in SAM optimization, this weakness of $\ell(\vw^{p_N})$ might hinder SAM-$N$ to converge an optimal solution.

\subsection{Simple Modification of Multi-step Ascent SAM}\label{sec:simple}

From the above observations, we presume that the inefficiency of multi-step ascent SAM is caused by the limitation of the gradient $\nabla\ell(\vw^{p_N})$, which has difficulty in minimizing $\ell(\vw)$ and $\ell(\vw^{p_1})$.
To address this issue, we suggest a simple modification that utilizes other gradients while preserving the information of $\nabla\ell(\vw^{p_N})$. Specifically, we force SAM-$N$ to use all gradients of intermediate perturbed weights $\vw^{p^n_N}$ as update directions. This modified update process can be formalized as follows:
\begin{equation}
    \vw_{t+1} = \vw_t - \eta \cdot \frac{1}{N} \sum_{n=1}^{N} \nabla \ell(\vw^{p^n_N}_t).
    \label{eq:msam_descent}
\end{equation}

By doing this, the modified SAM can address the limitation of $\nabla\ell(\vw^{p_N})$ through the use of $\nabla\ell(\vw^{p^n_N})$, which offers the advantage of diverse ascent directions. Moreover, this update procedure does not requires any additional forward and backward computations compared to the multi-step ascent SAM, since $\ell(\vw^{p^n_N})$ have already been acquired during the multi-step ascent. Although the proposed modification is simple and cheap, we can also consider other possible gradients, such as $\nabla\ell(\vw^{p_1})$ instead of $\nabla\ell(\vw^{p^1_N})$, to mitigate the inefficiency of multi-step ascent SAM. While alternative gradients can bring some benefits, the proposed modification show stable performance at the lowest cost. Further details are provided in Appendix.

The modified version of multi-step ascent SAM differs from the original SAM in that it can be viewed as minimizing multiple perturbed weights, i.e., $\sum \vw^{p^n_N}$ instead of $\vw^{p_N}$. To justify this modification in terms of the PAC-Bayesian Theorem, we here establish a theoretical analysis to demonstrate that it also provides an upper bound on generalization.

\begin{theorem} (PAC-Bayesian Theorem \cite{mcallester1999pac, dziugaite2017computing})\label{thm:pac}
For given training dataset $\mathcal{S}$ drawn from population distribution $\mathcal{D}$, define the training loss $L_{\mathcal{S}}(\boldsymbol{w}) := \frac{1}{|\mathcal{S}|}\sum_{\vx\in\mathcal{S}}\ell(\boldsymbol{w},\vx)$ and the population loss  $L_{\mathcal{D}}(\boldsymbol{w}) := \mathbb{E}_{\vx\sim\mathcal{D}}[\ell(\boldsymbol{w},\vx)]$.
With $1-a$ probability, for any prior $\mathcal{P}$ and posterior $\mathcal{Q}$, the following inequality holds:
\begin{align}\label{eq:pac}
    \mathbb{E}_{\boldsymbol{w} \sim \mathcal{Q}}\left[L_{\mathcal{D}}(\boldsymbol{w})\right] \leq \mathbb{E}_{\boldsymbol{w} \sim \mathcal{Q}}\left[L_{\mathcal{S}}(\boldsymbol{w})\right]+\sqrt{\frac{KL(\mathcal{Q} \| \mathcal{P})+\log \frac{|\mathcal{S}|}{a}}{2(|\mathcal{S}|-1)}}
\end{align}
where $KL(\cdot \| \cdot)$ is Kullback–Leibler divergence.
\end{theorem}
\begin{corollary}\label{cor:upperthm} (Stated informally)
Suppose $L_{\mathcal{D}}(\vw) \leq \mathbb{E}_{\vdelta \sim \mathcal{N}(\mathbf{0}, \sigma^2 I)}[L_{\mathcal{D}}(\vw+\vdelta)]$, where $\vdelta \in \mathbb{R}^k$ and $k$ is the number of parameters. Then, for any $0\leq\lambda\leq 1$, the following inequality holds with high probability
\begin{align}\label{eq:ours_upper_bound}
 L_{\mathcal{D}}(\boldsymbol{w}) &\leq 
\lambda\max _{\|\boldsymbol{\delta}\| \leq \rho_1} L_{\mathcal{S}}(\boldsymbol{w}+\boldsymbol{\delta})
 + (1-\lambda) \max _{\|\boldsymbol{\delta}\| \leq \rho_2} L_{\mathcal{S}}(\boldsymbol{w}+\boldsymbol{\delta})\\
 &+\sqrt{\frac{KL(\mathcal{Q}\|\mathcal{P}) + \log\frac{|\mathcal{S}|}{a}}{2(|\mathcal{S}|-1)}}
\end{align}
\end{corollary}

\begin{table*}[t!]
\centering
\caption{(CIFAR-10) Generalization performance results (\%). The bold number denotes the best performance for each column.}
\label{tab:cifar10}
\setlength{\tabcolsep}{1.05em}
\begin{tabular}{l|c|ccc|
>{\columncolor[HTML]{E5FFE5}}c}
\hhline{======}
\textbf{Ascent steps} & \textbf{Method}                                           & \textbf{Acc(A)} & \textbf{Acc(B)} & \textbf{Reported} & \cellcolor[HTML]{E5FFE5}{\textbf{Best}}  \\ \hhline{======}
\textbf{None} &\text{SGD}                             & 96.26±0.03                 & \multicolumn{1}{c}{-}      & -                              & 96.26±0.03                                       \\ \hline
\textbf{Single-step} &\text{SAM}   & 96.54±0.07                 &  96.38±0.14                & 96.52±0.13 & 96.54±0.07                                       \\
&\text{ESAM}    & {96.56±0.05}                 &  96.43±0.11                & 96.56±0.08 & {96.56±0.05}                                        \\
&\text{GSAM} & 96.05±0.05                 & 96.19±0.03                 & -                              & 96.19±0.03                                       \\
&\text{ASAM}        & 95.19±0.08                 & {96.53±0.09}                 & -                              & 96.53±0.09                                       \\ \hline
\textbf{Multi-step} &\text{SAM-2}                     & 96.58±0.13                 & 96.38±0.02      & -                              & 96.58±0.13                                      \\
&\text{MSAM-2}                     & \textbf{96.88±0.03}                 & \textbf{96.88±0.03}$^\dagger$      & -                              & \textbf{96.88±0.03}                                       \\ \hhline{======}
\multicolumn{6}{l}{$^\dagger$\footnotesize{MSAM shows the best performance under the setting where SAM shows the best performance.}}

\end{tabular}
\end{table*}

\begin{table*}[t!]
\centering
\caption{(CIFAR-100) Generalization performance results (\%). The bold number denotes the best performance for each column.}
\label{tab:cifar100}
\setlength{\tabcolsep}{1.05em}
\begin{tabular}{l|c|ccc
|>{\columncolor[HTML]{E5FFE5}}c}
\hhline{======}
\textbf{Ascent steps}  & \textbf{Method}                                           & {\textbf{Acc(A)}} & {\textbf{Acc(B)}} & {\textbf{Reported}} & {\cellcolor[HTML]{E5FFE5}{\textbf{Best}}}  \\ \hhline{======}
\textbf{None} & \text{SGD}                                              & 82.94±0.13                          & -                                   & -                 &  82.94±0.13             \\ \hline
\textbf{Single-step}& \text{SAM}   & {85.44±0.07}                          &  84.05±0.17                         & 85.10±0.20        & {85.44±0.07}    \\
&\text{ESAM}    & 78.87±0.20                          &  84.07±0.15                         & 84.51±0.01        & 84.51±0.01    \\
&\text{GSAM} & 83.36±0.39                          & 82.76±0.20                          & -                 & 83.36±0.39    \\
&\text{ASAM}        & 84.09±0.08                          &  \underline{84.36±0.15}                         & 83.68±0.12        & 84.36±0.15    \\ \hline
\textbf{Multi-step}& \text{SAM-2}                     & 84.91±0.08                          & 84.71±0.06                                   & -                 & 84.91±0.08    \\
&\text{MSAM-2}                     & \textbf{85.87±0.38}                          & \textbf{86.06±0.07}                                   & -                 & \textbf{86.06±0.07}    \\ \hhline{======}

\end{tabular}
\end{table*}

Proofs are presented in Appendix. \Coref{cor:upperthm} implies that utilizing multiple gradients of perturbed weights can also provide an upper bound on generalization.

\section{Experiments}\label{sec:experiments}

In this section, we experimentally demonstrate that the proposed modification can mitigate the inefficiency of multi-step ascent SAM and improve its performance. We conduct the experiments on CIFAR-10 \cite{krizhevsky2009learning}, CIFAR-100 \cite{krizhevsky2009learning}, and ImageNet \cite{imagnet}. For CIFAR-10 and CIFAR-100, we trained ResNet18 \cite{he2016deep} and Wide-ResNet-28-10 \cite{BMVC2016_87}, respectively. For ImageNet, we trained ResNet-50 \cite{he2016deep} with $224 \times 224$ resized images. The detailed settings are provided in the Appendix.

\subsection{Generalization performance}\label{sec:gen}
First, we compare the performance of multi-step ascent SAM and the Modified version of multi-step ascent SAM (MSAM). To conduct a comprehensive set of experiments, we also consider other variations of SAM from other studies, such as ESAM \cite{du2021efficient}, GSAM \cite{zhuang2021surrogate}, and ASAM \cite{kwon2021asam}.

\begin{table}[t]
\centering
\caption{(ImageNet) Generalization performance results (\%). The bold number denotes the best performance.}
\label{tab:imagenet}
\setlength{\tabcolsep}{1.05em}
\begin{tabular}{l|c|>{\columncolor[HTML]{E5FFE5}}c}
\hhline{===}
\textbf{Ascent steps} &\textbf{Method} & \textbf{ Acc } \\ \hhline{===}
\textbf{None} &SGD     & 76.0                                                 \\ \hline
\textbf{Single-step} &SAM   & 76.9                                                  \\
&ESAM* & 77.1                                                  \\
&GSAM*  & {77.2}                                                   \\
&ASAM* & 76.6                                               \\ \hline
\textbf{Multi-step} &SAM-2   & 77.1                                                   \\ 
&MSAM-2   & \textbf{77.9}                                                   \\ \hhline{===}      
\end{tabular} %
\\ *\small{The values were referred to each paper.} \quad\quad\quad
\end{table}

\paragraph{CIFAR-10 and CIFAR-100}

For the CIFAR datasets, we observe a wide range of $\rho$ values used in other studies. To ensure a fair comparison, we report three accuracy values on CIFAR: (1) \textit{Acc(A)}: Accuracy under the same $\rho$ and training setting where SAM achieves the best performance; (2) \textit{Acc(B)}: Accuracy under the radius $\rho$ proposed by authors, but with other settings remaining the same; (3) \textit{Reported}: Reported accuracy in their original paper. We provide the average accuracy and standard deviation across three different random seeds.

As shown in \Tabref{tab:cifar10} and \ref{tab:cifar100}, SAM-$2$ does not exhibit superior performance compared to other single-step ascent SAMs. However, in contrast, MSAM shows better performance than any other single-step ascent and SAM-$2$ for both CIFAR-10 and CIFAR-100.

\paragraph{ImageNet}
Similar to CIFAR-10 and CIFAR-100, SAM-$2$ shows similar or worse performance than other single-step ascent SAMs. However, MSAM effectively enhances the accuracy of SAM-$2$, leading to the highest performance among the variations of SAM. Thus, we can conclude that our modification of multi-step ascent SAM is beneficial.

\begin{figure*}[t]
\centering
    \includegraphics[width=0.98\linewidth]{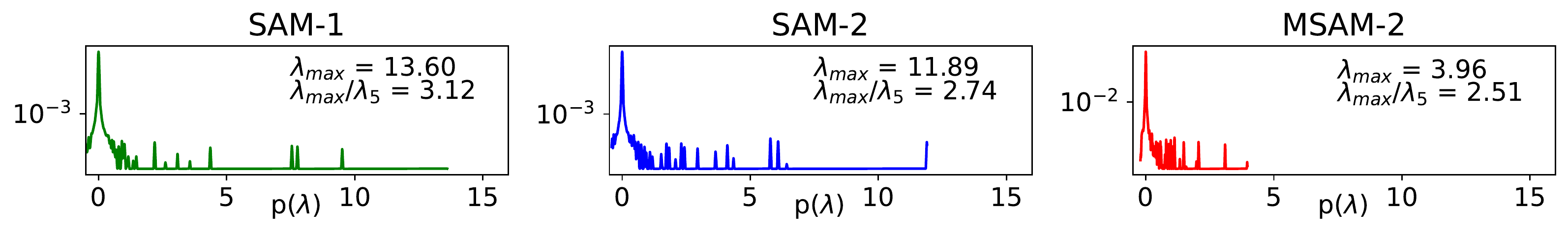} 
  \caption{(CIFAR-10) Spectrum of the Hessian $\mathbf{H}$ of the model trained by each method.}
  \label{fig:hessian}
\end{figure*}

\begin{table}[t]
  \begin{center}
    {\captionsetup{aboveskip=-10pt}
        {\renewcommand\arraystretch{1.0}
        \caption{Accuracy (\%) varying the number of ascent steps $N$.}
        \resizebox{0.4\textwidth}{!}{%
        \centering
        \begin{tabular}{c|cccc}
        \hhline{=====}
        Dataset    &  3        & 4   &  5   \\ \hline
        CIFAR-10 & 96.84\tiny{±0.10} & 96.76\tiny{±0.07} & 96.85\tiny{±0.07} \\
        CIFAR-100 & 86.05\tiny{±0.05} & 85.96\tiny{±0.17} & 85.89\tiny{±0.12} \\
        \hhline{=====}
        \end{tabular}%
        }
        \label{tab:sensitivity_main_n}
    }
    }
 \end{center}
\end{table}

\begin{table}[t]
  \begin{center}
    {\captionsetup{aboveskip=-10pt}
        {\renewcommand\arraystretch{1.0}
        \caption{Accuracy (\%) varying the step-size ratio $\rho_1:\rho_2$.}
        \begin{tabular}{c|ccc}
        \hhline{====}
        Dataset    & 1:2        & 1:1        & 2:1        \\ \hline
        CIFAR-10 & 96.88\tiny{±0.03} & 96.80\tiny{±0.08} & 96.86\tiny{±0.11} \\
        CIFAR-100 & 85.87\tiny{±0.38} & 85.69\tiny{±0.10} & 85.73\tiny{±0.13} \\ \hhline{====}
        \end{tabular}
        \label{tab:sensitivity_main_rhon}
        }
    }
 \end{center}
\end{table} 

\begin{figure}[t]
\centering
    \includegraphics[width=0.9\linewidth]{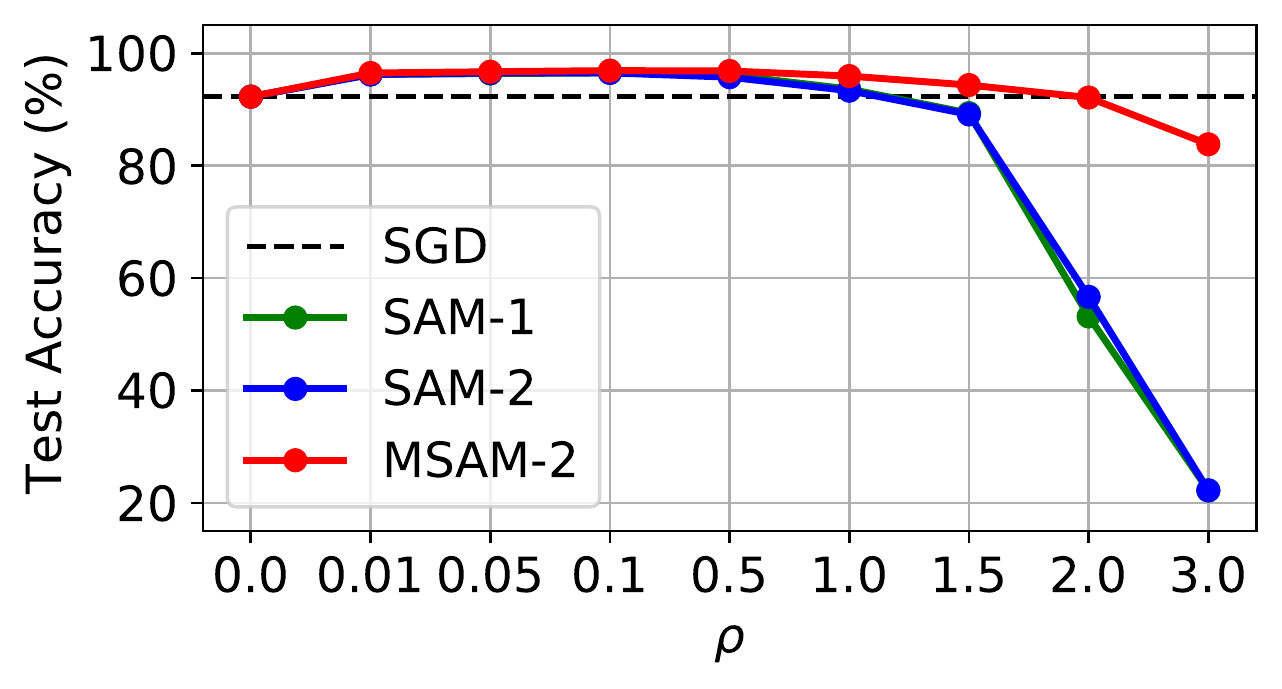} 
  \caption{(CIFAR-10) Sensitivity analysis on $\rho$.}
  \label{fig:diff_rho_msam}
\end{figure}

\subsection{Sensitivity Analysis}
The proposed modification of multi-step ascent SAM has two major hyper-parameters, the number of ascent steps $N$ and the radii $\rho_n$ during ascent steps. To gain a deeper understanding on the proposed modification and SAM, we conduct a sensitivity analysis on both hyper-parameters.

First, we compare the performance of SAM and MSAM for different $N$ under the step-size $\rho_n=\rho/N$. As previously observed in \Figref{fig:Barchart}, $N$ rarely affects the performance of SAM. In contrast, MSAM achieves higher performance than SAM for all $N\geq3$, indicating that using the gradient information of multiple perturbed weights is beneficial. The lack of improvement for $N\geq4$ may be related to \Figref{fig:cosine}, which shows that the cosine similarity between $\nabla\ell(\vw^{p_3})$ and $\nabla\ell(\vw^{p_4})$ is exactly one.  Although there might exist a proper number of ascent steps for specific cases, we leave further research on this possibility as future work.

Additionally, we conduct a sensitivity analysis on the step-size $\rho_n$ during ascent steps. For simplicity, in this experiment, we fix $N=2$ and $\rho=0.1$. As shown in \Tabref{tab:sensitivity_main_rhon}, MSAM exhibits a stable performance across all ratios. To push further, we compare SAM and MSAM with varying $\rho$ values ranging from 0.0 to 3.0 that is notably higher than the default value. The average accuracy for three random seeds is illustrated in \Figref{fig:diff_rho_msam}. Interestingly, even for larger $\rho$ values, MSAM shows 
stable performance than SAM-$1$ and SAM-$2$. We believe that this is an additional benefit of utilizing multiple gradients, each of which offers its own advantages.

\subsection{Better sharpness and Hessian spectra}

In order to verify that the proposed method enables the model to reach flatter minima, we calculate the Hessian spectra with their eigenvalues in \Figref{fig:hessian}. Here, we use the power iteration with 100 iterations. The model trained by MSAM has the smallest curvature, as demonstrated in the general distribution of eigenvalues. Furthermore, MSAM attains the lowest maximum eigenvalue and spectrum of the Hessian compared to SAM-$1$ and SAM-$2$.

\section{Conclusion}\label{sec:conclusion}
Through our analysis of the effect of the number of ascent steps on SAM optimization, we demonstrated that varying the number of ascent steps has a significant impact on the perturbed weights and their losses, and thus models trained with single and multiple ascent steps have distinct loss landscapes. Our findings and proposed modification discover that multi-step ascent can improve the performance of SAM, contrary to the previous belief. We believe that our study provides valuable insights into the optimization of SAM and opens up new opportunities for further research in the field of sharpness-aware training methods.





\bibliography{example_paper}
\bibliographystyle{icml2023}

\end{document}